\documentclass[acmtog]{acmart}

\usepackage{booktabs} 
\usepackage[ruled]{algorithm2e} 
\usepackage{makecell}

\acmDOI{}

\renewcommand\footnotetextcopyrightpermission[1]{}
\makeatletter
\renewcommand\@authorsaddresses{}
\makeatother

\setcitestyle{square}

\begin{document}
\acmSubmissionID{362} 

\title{Dynamic Temporal Alignment of Speech to Lips}

\author{Tavi Halperin}
\authornote{Indicates equal contribution}
\author{Ariel Ephrat}
\authornotemark[1]
\author{Shmuel Peleg}
\affiliation{%
  \institution{The Hebrew University of Jerusalem}}
\authorsaddresses{}
\renewcommand{\shortauthors}{Halperin, Ephrat, and Peleg}

\begin{abstract}
Many speech segments in movies are re-recorded in a studio during post-production, to compensate for poor sound quality as recorded on location. Manual alignment of the newly-recorded speech with the original lip movements is a tedious task. We present an audio-to-video alignment method for automating speech to lips alignment, stretching and compressing the audio signal to match the lip movements. This alignment is based on deep audio-visual features, mapping the lips video and the speech signal to a shared representation. Using this shared representation we compute the lip-sync error between every short speech period and every video frame, followed by the determination of the optimal corresponding frame for each short sound period over the entire video clip.   We demonstrate successful alignment both quantitatively, using a human perception-inspired metric, as well as qualitatively. The strongest advantage of our audio-to-video approach is in cases where the original voice in unclear, and where a constant shift of the sound can not give a perfect alignment. In these cases state-of-the-art methods will fail. 

\end{abstract}


\begin{teaserfigure}
  \centering
  \includegraphics[width=0.7\linewidth]{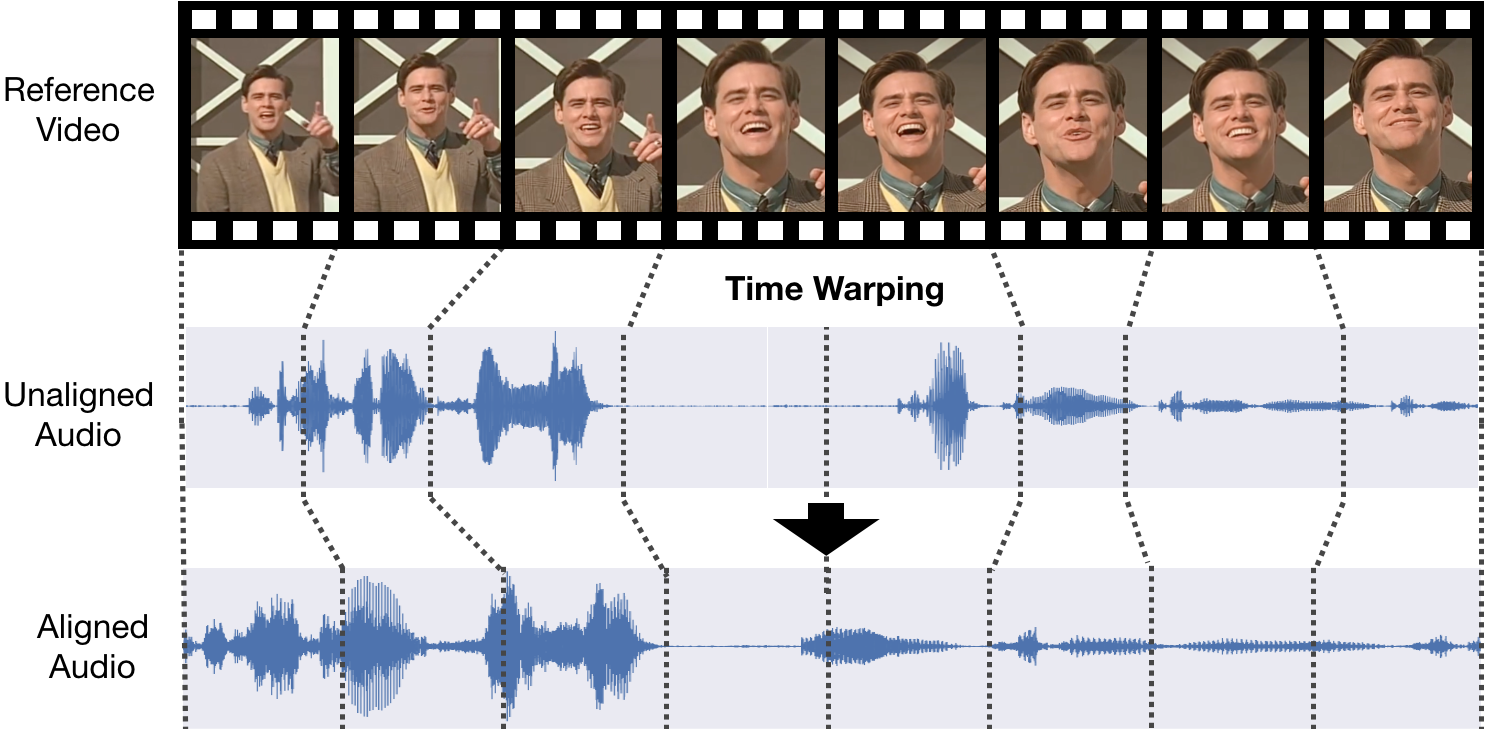}
  \caption{Given a speech video and a segment of corresponding, but unaligned, audio, we align the audio to match the lip movements in the video.}
\label{fig:teaser}
\end{teaserfigure}

\maketitle

\section{Introduction}
\label{sec:intro}

In movie filming, poor sound quality is very common for speech recorded on location.  Maybe a plane flew overhead, or the scene itself was too challenging to record high-quality audio. In these cases, the speech is re-recorded in a studio during post-production using a process called ``Automated Dialogue Replacement (ADR)'' or ``looping''. In ``looping'' the actor watches his or her original performance in a loop, and  re-performs each line to match the wording and lip movements.

ADR is a tedious process, and requires much time and effort by the actor, director, recording engineer, and the sound editor. One of the most challenging parts of ADR is aligning the newly-recorded audio to the actor's original lip movements, as viewers are very sensitive to audio-lip discrepancies. This alignment is especially difficult when the original on-set speech is unclear.

In this work we temporally align audio and video of a speaking person by using innovative deep audio-visual (AV) features that were suggested by \cite{chung2016out}. These features map the lips video and the speech signal to a shared representation. Unlike the original synchronization method of \citet{chung2016out}, which shifts the audio or the video clip by a global offset, we use dynamic temporal alignment, stretching and compressing the signal dynamically within a clip. This is usually a three-step process \cite{hosom2000automatic}: $(i)$ features are calculated for both the reference and the unaligned signals; $(ii)$ optimal alignment which maps between the two signals is found using dynamic time warping (DTW) \cite{rabiner1993fundamentals}; $(iii)$ a warped version of the unaligned signal is synthesized so that it temporally matches the reference signal \cite{ninness2008time}. In this paper we leverage the pre-trained AV features of \citet{chung2016out} to find an optimal audio-visual alignment, and then use dynamic time warping to obtain a new, temporally aligned speech video.

We demonstrate the benefits of our approach over a state-of-the-art audio-to-audio alignment method, and over \citet{chung2016out}, using a human perception-inspired quantitative metric. Research has shown that the detectability thresholds of lack of synchronization between audio and video is +45 ms when the audio leads the video and -125 ms when the audio is delayed relative to the video. The broadcasting industry uses these thresholds in their official broadcasting recommendations \cite{itu1359}. In order to evaluate the perceptive quality of our method's output, our quantitative error measure is therefore the percentage of aligned frames which are mapped outside the undetectable region, relative to ground truth alignment. It should be noted that comparison to an audio-to-audio alignment method can only be performed when a clear reference audio signal exists, which may not always be the case. In that scenario, dynamic audio-to-visual or visual-to-visual alignment is the only option, a task which, to the best of our knowledge, has not yet been addressed.

To summarize, our paper's main contribution is a method for fully automated dialogue replacement in videos (ADR). We leverage the strength of deep audio-visual speech synchronization features of \citet{chung2016out} and suggest a dynamic temporal alignment method. To the best of our knowledge, our paper is the first to propose a method for dynamic audio-to-visual time alignment. 

\section{Related work}
\label{sec:related}

We briefly review related work in the areas of audio and video synchronization and alignment, as well as speech-related video processing.

\paragraph{Audio-to-video synchronization.}

Audio-to-video synchronization (AV-sync), or \textit{lip-sync}, refers to the relative timing of auditory and visual parts of a video. Automatically determining the level of AV-sync in a video has been the subject of extensive study within the computer vision community over the years, as lack of synchronization is a common problem. In older work, such as \citet{lewis1991automated}, \emph{phonemes} (short units of speech) are recognized and subsequently associated with mouth positions to synchronize the two modalities. \citet{morishima2002audio} classifies parameters on the face into \emph{visemes} (short units of visual speech), and uses a viseme-to-phoneme mapping to calculate synchronization. \citet{zoric2005real} train a neural network to solve this problem.

In more recent work, methods have been proposed which attempt to find audio-visual correspondences without explicitly recognizing phonemes or visemes, such as \citet{bredin2007audiovisual} and \citet{sargin2007audiovisual} who use canonical correlation analysis (CCA). \citet{marcheret2015detecting} train a neural network-based classifier to determine the synchronization based on pre-defined visual features.
In a recent pioneering work \citet{chung2016out} have proposed a model called \emph{SyncNet}, which learns a joint embedding of visual face sequences and corresponding speech signal in a video by predicting whether a given pair of face sequence and speech track are synchronized or not. They show that the learned embeddings can be used to detect and correct lip-sync error in video to within human-detectable range with greater than 99\% accuracy. 

The common denominator of the above works is that they attempt to detect and correct a global lip-sync error, i.e. the global shift of the audio signal relative to the video. In this work, we leverage the audio-visual features of SyncNet to perform dynamic time alignment, which can stretch and compress very small units of the unaligned (video or audio) signal to match the reference signal.

\paragraph{Automatic time alignment of sequences.}
Dynamic time warping (DTW) \cite{sakoe1978dynamic} uses dynamic programming to find the optimal alignment mapping between two temporal signals by minimizing some pairwise distance (e.g. Euclidean, cosine, etc.) between sequence elements. This algorithm has been used extensively in the areas of speech processing \cite{sakoe1978dynamic,hosom2000automatic,king2012noise} and computer vision \cite{zhou2008aligned,gong2011dynamic,halperin2018egosampling} for e.g. temporal segmentation and frame sampling, among many scientific disciplines.

\citet{king2012noise} propose a new noise-robust audio feature for performing automatic audio-to-audio speech alignment using DTW. Their feature models speech and noise separately, leading to improved ADR performance when the reference signal is degraded by noise. This method of alignment essentially uses audio as a proxy for aligning the re-recorded audio with existing lip movements. When the reference audio is very similar to the original, this results in accurate synchronization. However, when the reference signal is significantly degraded (as a result of difficult original recording conditions), our method overcomes this problem per performing audio-to-video alignment directly, resulting in higher-quality synchronization. In addition, when a reference audio signal is unavailable, direct audio-to-video alignment is the only option.

\paragraph{Speech-driven video processing.}
There has been increased interest recently within the computer vision community in leveraging natural synchrony of simultaneously recorded video and speech for various tasks. These include audio-visual speech recognition \cite{Ngiam2011MultimodalDL,mroueh2015deep,feng2017audio}, predicting a speech signal or text from silent video (\emph{lipreading}) \cite{ephrat2017vid2speech,ephrat2017improved,Chung2016LipRS}, and audio-visual speech enhancement \cite{ephrat2018looking,afouras2018conversation,owens2018audio}.

A large and relevant body of audio-visual work is speech-driven facial animation, in which, given a speech signal as input, the task is to generate a face sequence which matches the input audio \cite{bregler1997video,cao2004real,chang2005transferable,furukawa2016video,taylor2017deep}. We do not attempt to provide a comprehensive survey of this area, but mention a few recent works. \citet{garrido2015vdub} propose a system for altering mouth motion of an actor in a video, so that it matches a new audio track containing a translation of the original audio (\emph{dubbing}. \citet{suwajanakorn2017synthesizing} use an RNN to map audio features to a 3D mesh of a specific person, and \citet{chung2017you} train a CNN which takes audio features and a still face frame as input, and generates subject-independent videos matching the audio. \citet{thies2016face2face} don't use audio explicitly, but propose a real-time system for reenacting the face movement of a source sequence on a target subject. While the above works succeed in producing impressive results, they require the subject to be in a relatively constrained setting. This is oftentimes not the case in difficult-to-record movie scenes which require ADR, where the goal is to align the video and audio without modifying the pixels in the video frames.

\section{Dynamic Alignment of Speech and Video}
\label{sec:method}

\begin{figure*}[th]
  \centering
  \includegraphics[width=\linewidth]{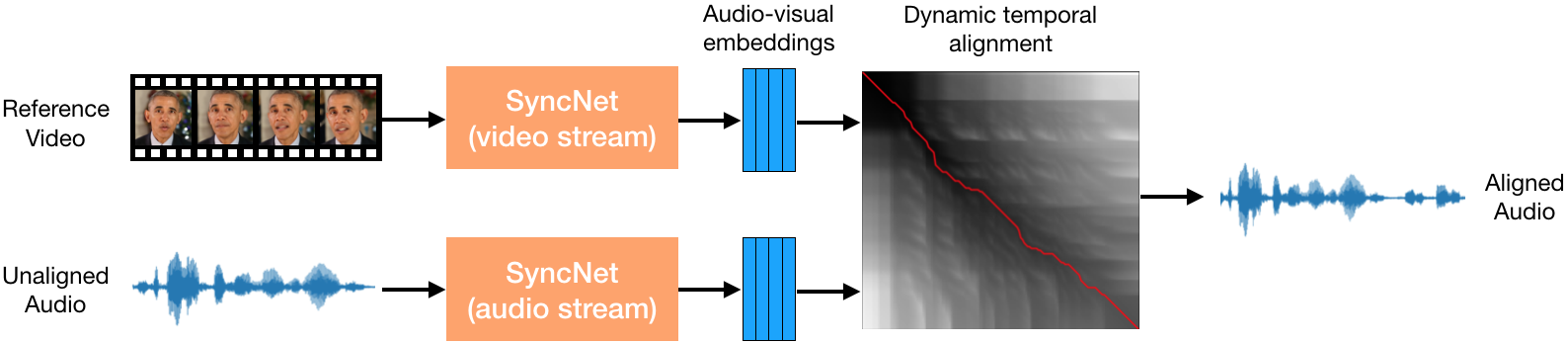}
  \caption{{\bf High-level diagram of our speech to lips alignment:} Given unaligned video and speech: (i) SyncNet features are computed for both; (ii) dynamic time warping is performed for optimal alignment between the features; (iii) A new speech is synthesized that is now aligned with the video.}
\label{fig:method}
\end{figure*}

Our speech to lips alignment is comprised of three main stages: audio-visual feature extraction, finding an optimal alignment which maps between audio and video, and synthesizing a warped version of the unaligned signal to temporally match the reference signal. An overview of our method is illustrated in Figure \ref{fig:method}.

\subsection{Audio-Visual Feature Extraction}
\label{ssec:features}

We use SyncNet~\cite{chung2016out} to extract language-independent and speaker-independent audio-visual embeddings. The network was trained to synchronize audio and video streams which were recorded simultaneously. This type of synchronization is termed `linear' as the audio is shifted by a constant time delta throughout the entire video. SyncNet encodes short sequences of 5 consecutive frames with total duration of 200 ms. or the equivalent amount of audio into a shared embedding space. We use the network weights provided by the authors, which were trained to minimize $l_2$ distance between embeddings of synchronized pairs of audio and video segments while maximizing distance between non matching pairs. We define the data term for our Dynamic Programing cost function to be pairwise distances of these embeddings. 

\subsection{Dynamic Time Warping for Audio-Visual Alignment}
\label{ssec:alignment}
Naturally, as the number of possible mouth motions is limited, there are multiple possible low cost matches for a given short sequence. For example, segments of silence in different parts of the video are close in embedding space. SyncNet solves this by averaging time shift prediction over the entire video. We, however, are interested in assigning per frame shifts, therefor we use dynamic time warping.

Our goal here is to find a mapping (`path') with highest similarity between two sequences of embeddings $A=(a_1,...,a_N), B=(b_1,...,b_M)$, subject to non decreasing time constraint: if the path contains $(a_i,b_j)$ then later frames $a_{i+k}$ may only match later audio segments $b_{j+l}$. Additional preferences are (i) audio delay is preferred over audio advance with respect to reference video (a consequence of the different perception of the two); (ii) smooth path, to generate high quality audio; (iii) computationally efficient. We will now describe how we meet these preferences. 

We solve for optimal path using Dijkstra's shortest path algorithm~\cite{dijkstra1959note}. We construct a data cost matrix $C$ as pairwise dot products between embeddings from the reference video and the embeddings of an unaligned audio. Each matrix element is associated with a graph node, and edges connect node $(i,j)$ to $\{(i+1,j), (i,j+1), (i+1,j+1)\}$ so that the non decreasing time constraint holds. 

Classically, the weight on an edge pointed at $(i,j)$ is the matrix value of the target element $C_{i,j}$. To better fit the perceptual attributes of consuming video and audio we modify the cost to prefer a slight delay by assigning the weight $0.5*C_{i,j}+0.25*C_{i-1,j}+0.25*C_{i-2,j}$. Relative improvement which stems from this modification is studied in Section \ref{sec:experiments}.

We assume the two modalities are cut roughly to the same start and end points, so we find a minimal path from $(0,0)$ to $(N,M)$. We experimented with looser constraint by adding quiet periods on start and end points, and did not find any significant difference in results. 

If other modalities exist, i.e reference audio and unaligned video, we compute 4 cross distances between embeddings of reference and unaligned, and assign the matrix element with the \emph{minimal} of all four. This helps mitigate effects of embedding noise from e.g face occlusion or sudden disrupting sounds. We found out that even in the absence of such noise, combining different modalities improves the alignment.    

In terms of our cost matrix, Syncnet's global shift corresponds to selecting the path as a diagonal on the matrix. 

\begin{figure}
  \centering
  \includegraphics[width=0.7\linewidth]{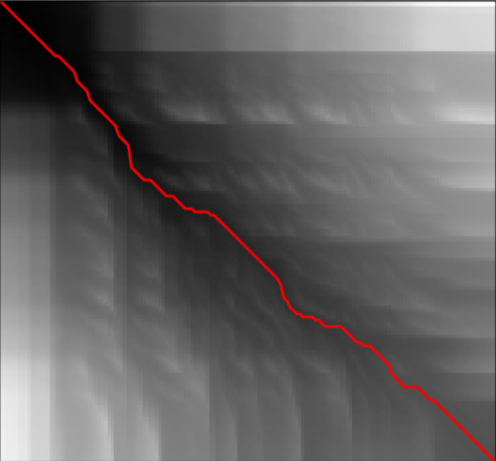}
  \caption{{\bf Cumulative cost matrix for dynamic programming:} The figure shows a sample matrix containing cumulative frame matching costs for a reference video and unaligned audio pair. Each matrix element contains the cumulative cost of matching an audio frame (row) and a video frame (column). Darker entries correspond to lower cost, and the optimal alignment is the path which minimizes the overall matching cost (shown here in red). Note that there are many similar structures because, for example, two different silent segments would have similar rows.}
\label{fig:dp_matrix}
\end{figure}

To avoid unnecessary computations, we only compute costs of nodes and edges in a strip around the `diagonal' $(0,0)\rightarrow(N,M)$. A sample full matrix is shown in Figure~\ref{fig:dp_matrix} for visualization.

\subsection{Smoothing the Path}
While the optimal path between sequences of embeddings is found, the quality of the generated audio based on that path may be degraded due to strong accelerations in the alignment. We first smooth the path with a Laplacian filter, then with a Gaussian. The amount of smoothing is chosen adaptively so that the smoothed path will not deviate from the original by more than a predefined value $\lambda$. Usually we set $\lambda<0.1$ seconds, well within the boundaries of undetectable misalignment. This value may be changed for signals with specific characteristics or for artistic needs. After smoothing, the path is no longer integer valued, and interpolation is needed for voice synthesis. 

\subsection{Synthesis of New Signal}
\label{ssec:synthesis}
In standard ADR, the task is to warp the audio without modifying the original video frames. Therefore, we use the alignment to guide a standard audio warp method. We use a fairly simple phase vocoder ~\cite{laroche1999new} to stretch and compress the audio stream according to the alignment, without affecting the pitch. 
This method uses the short-time Fourier transform (STFT) computed over the audio signal. We used audio sampled at 16KHz, each STFT bin (including complex values for each time-frequency cell) was computed on a window of size 512 audio samples, with $1/2$ window overlap between consecutive STFT bins.
The STFT magnitude is time warped, and phases are fixed to maintain phase differences between consecutive STFT windows. Since our alignment is based on video frames, its accuracy is only at time steps of 40ms, while the time step between STFT bins 16 ms. We create the alignment between  STFT bin by re-sampling the frame-level alignment.


\section{Experiments and results}
\label{sec:experiments}

Our main motivating application is fully automating the process of dialogue replacement in movies, such that the re-recorded speech can be combined with the original video footage in a seamless manner. We tested our method, both quantitatively and qualitatively, in a variety of scenarios.

\paragraph{Evaluation.}
Quantitative evaluation was performed using a human perception-inspired metric, based on the maximum acceptable audio-visual asynchrony used in the broadcasting industry. According to the International Telecommunications Union (ITU), the auditory signal should not lag by more than 125 ms or lead by more than 45 ms. Therefore, the error metric we use is the percentage of frames in the aligned signal which fall outside of the above acceptable range, compared to the ground truth alignment.

\subsection{Alignment of Dually-Recorded Sentences}
\label{ssec:dually}

In this task, given a sentence recorded twice by the same person---one \emph{reference} signal, and the other \emph{unaligned}---the goal is to find the optimal mapping between the two, and warp the unaligned audio such that it becomes aligned with the reference video.

To our knowledge, there are no publicly available audio-visual datasets containing this kind of dually-recorded sentences, which are necessary for evaluating our method. To this end, we collected recordings of the same two sentences (\emph{sa1} and \emph{sa2} from the TIMIT dataset \cite{timit}) made by four male speakers and one female speaker. The only instruction given to the speakers was to speak naturally. Therefore, the differences in pitch and timing between the recordings were noticeable, but not extremely distinct.

The dataset for this experiment was generated by mixing the original \emph{unaligned} recordings with two types of noise, at varying signal-to-noise (SNR) levels. The types of noise we used, \emph{crowd} and \emph{wind}, are characteristic of interferences in indoor and outdoor recording environments, respectively. In order to demonstrate the effectiveness of our approach in noisy scenarios, we generated noise at three different levels: $0$, $-5$, and $-10$ dB.

Alignment of each segment is performed using the following dynamic programming setups: ($a$) Alignment of unaligned audio to reference video (\emph{Audio-to-video}); ($b$) Audio-to-video alignment with the additional delay constraint detailed in Section~\ref{ssec:alignment} (\emph{Audio-to-video + delay}); ($c$) All combinations of modality-to-modality alignment, namely, audio-to-audio, audio-to-video, video-to-audio, and video-to-video, taking the step with minimum cost at each timestep (\emph{All combinations}); ($d$) All modality combinations, with the additional delay constraint (\emph{All combinations + delay}).

We compare our method to the state-of-the-art audio-to-audio alignment method of \citet{king2012noise}, which has been implemented as the \emph{Automatic Speech Alignment} feature in the Adobe Audition digital audio editing software \cite{adobe}. This method uses noise-robust features as input to a dynamic time warping algorithm, and obtains good results when the reference signal is not badly degraded. As a baseline, we also compare to the method of \citet{chung2016out} for finding a global offset between signals, whose audio-visual features we use as input to our method.

Since we have no ground truth mapping between each pair of recorded sentences, we adopt the method used by \citet{king2012noise} for calculating a ``ground truth'' alignment. They use conventional Mel-Frequency Cepstral Coefficients (MFCCs) to calculate alignment between reference and unaligned audio clips, with no noise added to the reference. Time-aligned synthesized ``ground truth'' signals were manually verified to be satisfactory, by checking audio-visual synchronization and comparing spectrograms.


\begin{table*}
\small
\centering
\caption{{\bf Quantitative analysis and comparison with prior art:}  This table shows the superiority of our approach over (i) a state-of-the-art audio-to-audio alignment method, implemented as feature in Adobe Audition \cite{king2012noise}, and (ii) SyncNet \cite{chung2016out}. The error is expressed as percentage of aligned frames outside the undetectable asynchrony range (-125 to +45 ms). The results demonstrate that even at lower noise levels, our Audio-to-Video and our combined modality (Audio to Video+Audio) approaches have improved performance over existing methods. At extremely high noise levels, our method has a clear and significant advantage. The delay is described in Sec.~\ref{ssec:alignment}.
}
\begin{tabular}{lcccccc}
\toprule[1.5pt]
& \multicolumn{3}{c}{\bf ``Crowd'' noise} & \multicolumn{3}{c}{\bf ``Wind'' noise} \\
& 0 dB & -5 dB & -10 dB & 0 dB & -5 dB & -10 dB \\
\midrule
SyncNet \cite{chung2016out} &  88.49 & 88.49 & 88.49 & 88.49 & 88.49 & 88.49 \\
Adobe Audition \cite{king2012noise} & 4.07 & 10.23 & 10.61 & 4.85 & 4.93 & 10.09  \\
\midrule
Audio-to-Video  & 7.26 & 7.26 & 7.26  & 7.26 & 7.26 & 7.26    \\
Audio-to-Video (with delay) 		  & 4.12 & 4.12 & 4.12 & 4.12 & 4.12 & 4.12     \\
Audio to Video+Audio  & 2.03 & 1.98 & \bf 2.03 & 3.77 & 5.04 & 5.85  \\
Audio to Video+Audio (with delay) & \bf 0.61 & \bf 0.88 & 4.25 & \bf 1.21 & \bf 1.22 & \bf 4.03      \\
\bottomrule[1.5pt] \\
\end{tabular}
\label{tb:comparison}
\end{table*}

Table~\ref{tb:comparison} shows the superiority of our approach, with the error expressed as percentage of aligned frames outside the undetectable asynchrony range (-125 to +45 ms). The results demonstrate that at even at lower noise levels, our AV and combined modality approaches give improved performance over existing methods. At extreme noise levels, e.g crowd noise at -10 dB, our combined method has an average of only around 2\% of frames outside the undetectable region, whereas the method of \citet{king2012noise} has over 10.5\%. Alignment using SyncNet results in 88\% of the frames outside the undetectable region, has it attempts to find an optimal global offset.

\begin{figure}
  \centering
  \includegraphics[width=\linewidth]{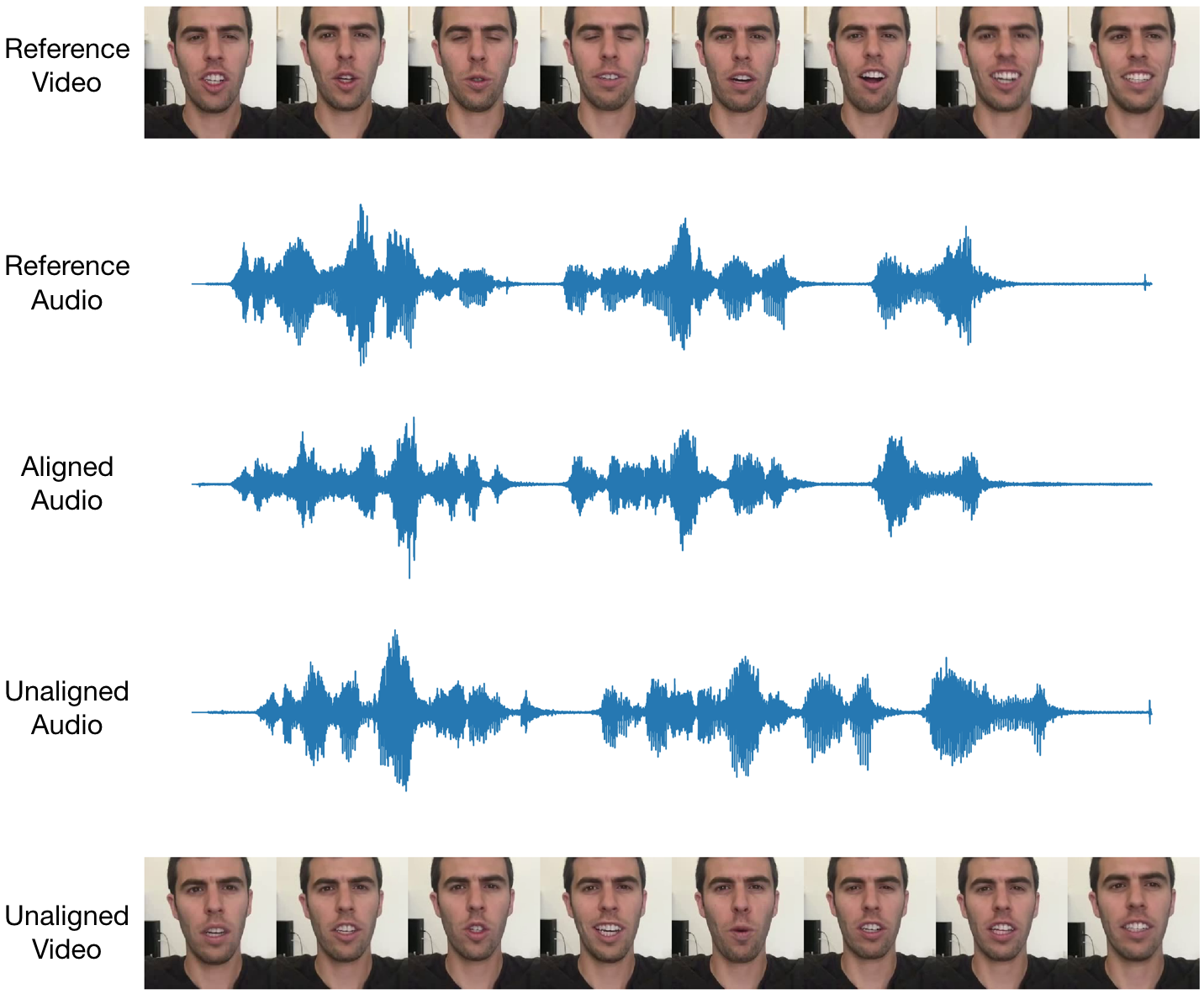}
  \caption{{\bf Example of reference and unaligned waveforms:} This figure shows examples of reference, unaligned and aligned video and audio waveforms for one of the dually-recorded sentences in our dataset.}
\label{fig:waveforms}
\end{figure}

Figure~\ref{fig:waveforms} shows examples of reference, unaligned and aligned video and audio waveforms for one of the dually-recorded sentences in our dataset. The videos of this example can be viewed in our supplementary video.

\subsection{Alignment of Synthetically Warped Sentences}
\label{ssec:synthetic}

In this task, we set out to investigate the limits of our method, in terms of degradation of both the audio and video parts of the reference signal as well as optimal segment duration for alignment. To this end, we use segments from a dataset containing weekly addresses given by former president Barack Obama, which are synthetically warped using mappings obtained from the dataset we created for the previous experiment. These mappings are representative of the natural variation in pronunciation when people record the same sentence twice. The goal in this experiment is to find the optimal alignment between the original reference video and the synthetically warped video.

\paragraph{Robustness to signal degradation}
In order to test the robustness of our method to various forms of degraded reference signals, we start with 100 same-length segments from the Obama dataset, and degrade the reference signals in following ways: ($i$) by adding crowd noise at $-10$ dB to each reference audio signal; ($ii$) by silencing a random one-second segment of each reference audio signal; ($iii$) by occluding a random one-second segment of each reference video sequence with a black frame; ($iv$) by combining random silencing and random occlusions ($ii$ + $iii$).

Each reference degradation is tested using the dynamic programming setups used in the previous experiment, namely: \emph{Audio-to-video}, \emph{Audio-to-video + delay}, \emph{All combinations}, and \emph{All combinations + delay}. Here too, we compare to the global offset method of \citet{chung2016out}, and add the error percentage of frames in the \emph{Unaligned} signal as a baseline.

\begin{table*}
\small
\centering
\caption{{\bf Analysis of robustness to degraded reference signals:} Alignment performance when the reference signal has undergone several types of degradation: (i) High noise, (ii) random 2-second silence in audio and (iii) 2-second blackout of video frames. The  error is expressed as percentage of aligned frames outside the undetectable asynchrony range. Note that using SyncNet \cite{chung2016out}, performing just a time shift, resukted in more errors than even the unaligned sound.}
\begin{tabular}{lcccc}
\toprule[1.5pt]
& \thead{\bf Crowd \\ \bf noise (-10dB)} & \thead{\bf Random \\ \bf silence} & \thead{\bf Random \\ \bf occlusion}  &\bf \thead{\bf Silence + \\ \bf occlusions} \\
\midrule
Unaligned Voice         & 33.77    & 34.03    & 37.87    &  32.73  \\
SyncNet \cite{chung2016out} & 71.37 & 73.34 & 78.74 & 84.0 \\
\midrule
Audio-to-Video 				 & \bf 2.63 & 2.92     & 16.16    & 15.17    \\
Audio-to-Video (with delay)		 & 3.35     & \bf 2.62 & 14.17     & 9.76     \\
Audio to Video+Audio  & 2.83     & 2.71     & \bf 3.08 & 5.78   \\
Audio to Video+Audio (with delay) & 5.45     & 3.14     & 4.12     & \bf 5.04      \\
\bottomrule[1.5pt] \\
\end{tabular}
\label{tb:synthetic_comp}
\end{table*}

Table~\ref{tb:synthetic_comp} shows the results of this experiment. When the audio is severely degraded with either loud noise or random silence, performing direct audio-to-video alignment performs best. When the reference video signal is degraded with occlusions, our method relies more on the audio signal, and combining both the audio and video of the reference video works best. Example videos of degraded reference video and the resulting alignment can be viewed in our supplementary video.

\paragraph{Effect of segment duration}
In order to investigate the effect segment duration has on alignment performance, we performed alignment on 100 segments from the Obama dataset of various durations between 3 to 15 seconds. There was no clear trend in the results of this study, leading us conclude that segment duration (within the aforementioned range) has a negligible effect on the performance of our method.

\subsection{Alignment of Two Different Speakers}
\label{ssec:different}
While not the main focus of our work, various additional alignment scenarios can be addressed using our audio-visual alignment method. One of these is alignment of two different speakers.

\begin{figure}
  \centering
  \includegraphics[width=.9\linewidth]{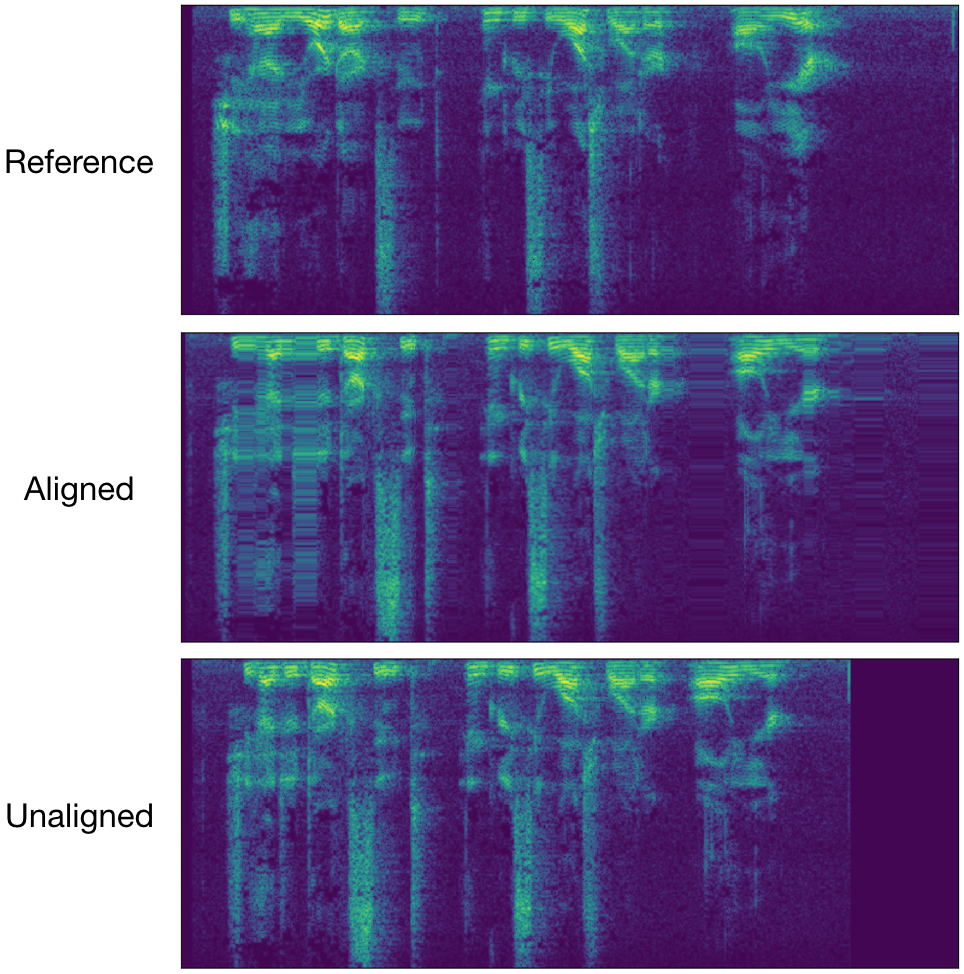}
  \caption{{\bf Spectrograms of alignment of different speakers:} This figure shows spectrograms of three signals: speech of one speaker used as reference (top); speech of a different speaker who we would like to align to the reference (bottom); aligned speech using our audio-to-video method (middle).}
\label{fig:warped_spec_different_speakers}
\end{figure}

Since audio and visual signals are mapped to a joint synchronization embedding space which, presumably, places little emphasis on the identity of the speaker, we can use our method to align of two different speakers saying the same text. For this task, we use videos from the TCD-TIMIT dataset \cite{harte2015tcd}, which consists of 60 volunteer speakers reciting various sentences from the TIMIT dataset \cite{timit}. We evaluated our results qualitatively, and included an example in our supplementary video, involving alignment between male and female subjects. Figure~\ref{fig:warped_spec_different_speakers} shows example spectrograms of reference, unaligned and aligned signals of two different speakers.

\section{Limitations and Conclusion}
\label{sec:conclusion}

Our method is currently limited by the quality of its synthesized speech, which is sometimes of poorer quality than original due to challenging warps. Also, in cases of clean reference speech, our method is comparable to existing audio-to-audio alignment.

In conclusion, a method was presented to align speech to lip movements in video using dynamic time warping. The alignment is based on deep features that map both the face in the video and the speech into a common embedding space. Our method makes it easy to create accurate Automated Dialogue Replacement (ADR), and have shown it to be superior to existing methods, both quantitatively and qualitatively. ADR is possible using speech of the original speaker, or even the speech of another person.

\bibliographystyle{ACM-Reference-Format}
\bibliography{references}
\end{document}